\documentclass[10pt,twocolumn,letterpaper]{article}

\usepackage{wacv}              % To produce the CAMERA-READY version
\usepackage{graphicx}
\usepackage{amsmath}
\usepackage{amssymb}
\usepackage{booktabs}
\usepackage{float}
\usepackage{algorithm}
\usepackage{algpseudocode}

\usepackage[pagebackref,breaklinks,colorlinks]{hyperref}
\usepackage[accsupp]{axessibility} % Improves PDF readability for those with disabilities.

\usepackage[capitalize]{cleveref}
\crefname{section}{Sec.}{Secs.}
\Crefname{section}{Section}{Sections}
\Crefname{table}{Table}{Tables}
\crefname{table}{Tab.}{Tabs.}

\def\wacvPaperID{2221} % *** Enter the WACV Paper ID here
\def\confName{WACV}
\def\confYear{2025}

\begin{document}

\title{A Conic Transformation Approach for Solving the Perspective-Three-Point Problem}

\author{
Haidong Wu\hspace{1cm}Snehal Bhayani\hspace{1cm}Janne Heikkilä\\
Center for Machine Vision and Signal Analysis\\
University of Oulu, Oulu, Finland\\
{\tt\small Haidong.Wu@oulu.fi, \tt\small Snehal.Bhayani@oulu.fi, \tt\small Janne.Heikkila@oulu.fi}
% Haidong Wu\\
% Center for Machine Vision and Signal Analysis\\
% University of Oulu, Oulu, Finland\\
% {\tt\small Haidong.Wu@oulu.fi}
% \and
% Snehal Bhayani\\
% Center for Machine Vision and Signal Analysis\\
% University of Oulu, Oulu, Finland\\
% {\tt\small Snehal.Bhayani@oulu.fi}
% \and
% Janne Heikkilä\\
% Center for Machine Vision and Signal Analysis\\
% University of Oulu, Oulu, Finland\\
% {\tt\small Janne.Heikkila@oulu.fi}
}
\maketitle

%%%%%%%%% ABSTRACT
\begin{abstract}
We propose a conic transformation method to solve the Perspective-Three-Point (P3P) problem. In contrast to the current state-of-the-art solvers, which formulate the P3P problem by intersecting two conics and constructing a degenerate conic to find the intersection, our approach builds upon a new formulation based on a transformation that maps the two conics to a new coordinate system, where one of the conics becomes a standard parabola in a canonical form. This enables expressing one variable in terms of the other variable, and as a consequence, substantially simplifies the problem of finding the conic intersection. Moreover, the polynomial coefficients are fast to compute, and we only need to determine the real-valued intersection points, which avoids the requirement of using computationally expensive complex arithmetic. 

While the current state-of-the-art methods reduce the conic intersection problem to solving a univariate cubic equation, our approach, despite resulting in a quartic equation, is still faster thanks to this new simplified formulation. Extensive evaluations demonstrate that our method achieves higher speed while maintaining robustness and stability comparable to state-of-the-art methods.

\end{abstract}

%%%%%%%%% BODY TEXT
\section{Introduction}
\label{sec:intro}
The Perspective-Three-Point (P3P) problem (\cref{fig:camera_model}) is a fundamental problem in geometric computer vision, which involves recovering the relative pose (including rotation and translation) between the camera and world coordinate systems from three pairs of 3D points and their corresponding 2D projections on the image plane.
Solutions to this problem are widely applied in fields such as augmented reality~\cite{marchand2015pose}, visual SLAM~\cite{mur2017orb}, photogrammetry~\cite{yuan1989general}, 
and robotics~\cite{abidi1995new}.

The P3P problem has a long history of development. In 1841, Grunert~\cite{grunert1841pothenotische} first demonstrated that the P3P problem can yield up to four feasible solutions. 
Since then, numerous methods have been introduced, with a selection ~\cite{grunert1841pothenotische, finstenvalder1937ruckwartseinschneiden, merritt1949explicit, fischler1981random, linnainmaa1988pose, grafarend1989dreidimensionaler}, reviewed and compared by Haralick \textit{et al.}~\cite{haralick1991analysis} in terms of numerical accuracy.
% Since then, a variety of approaches have been proposed to solve the problem. In 1991, Haralick \textit{et al}.~\cite{haralick1991analysis} offered a review of the major P3P algorithms that had been developed up to that time. 
% Gao \textit{et al}.~\cite{gao2003complete} used Wu-Ritt’s zero decomposition algorithm to provide the first complete triangular decomposition of the P3P equation system, leading to the first complete analytical solution to the P3P problem.
Gao \textit{et al.}~\cite{gao2003complete} applied Wu-Ritt’s zero decomposition algorithm ~\cite{wen1986basic} to achieve the first complete triangular decomposition of the P3P equation system, resulting in the first fully analytical solution to the P3P problem.
Kneip \textit{et al.}~\cite{kneip2011novel} and Masselli and Zell~\cite{masselli2014new} introduced a novel approach to solving the P3P problem by directly computing the absolute position and orientation of the camera, avoiding the computation of features' distances. Ke \textit{et al.}~\cite{ke2017efficient} directly determined the camera’s orientation by applying geometric constraints to construct a system of trigonometric equations. 
Banno~\cite{banno2018p3p} proposed a method to represent the rotation matrix as a function of the distances and Nakano \textit{et al.}~\cite{nakano2019simple} extended this approach with a simplified derivation that can be expressed in closed form.
These methods~\cite{gao2003complete,kneip2011novel,ke2017efficient,banno2018p3p,nakano2019simple} reduce the P3P problem to solving a quartic equation. Additionally, the P3P problem has also been simplified to solving a cubic equation in some approaches. Persson and Nordberg~\cite{Persson_2018_ECCV} proposed a method for solving the P3P problem by finding the single real root of a cubic equation, demonstrating significant effectiveness.
Ding \textit{et al.}~\cite{Ding_2023_CVPR} formulated the P3P problem as finding the intersection of two conics and solved it by determining the single real root of a cubic equation to construct a degenerate conic represented by two lines. To the best of our knowledge, the solver by Ding \textit{et al.}~\cite{Ding_2023_CVPR} achieves better numerical stability and higher speed compared to previous works.

In this paper, we address the P3P problem by deriving and solving a quartic equation. Following a similar strategy as in~\cite{Ding_2023_CVPR, Persson_2018_ECCV}, we first formulate the P3P problem as finding the intersections of two conics.
Specifically, the contributions are:

1. We propose a conic transformation that maps the two conics to a new coordinate system, where one of the conics becomes a standard parabola in a canonical form. The intersection of two conics in the new coordinate system is determined by the roots of a quartic equation, with coefficients that can be easily computed.

2. We propose a strategy to reduce the computational cost by determining the real transformation matrix and solving for the real roots of the quartic equation.

3. Extensive experiments demonstrate that our method achieves superior speed compared to state-of-the-art methods while maintaining comparable numerical stability and accuracy.

\section{Problem}
\begin{figure}[t]
  \centering
   \includegraphics[width=1\linewidth]{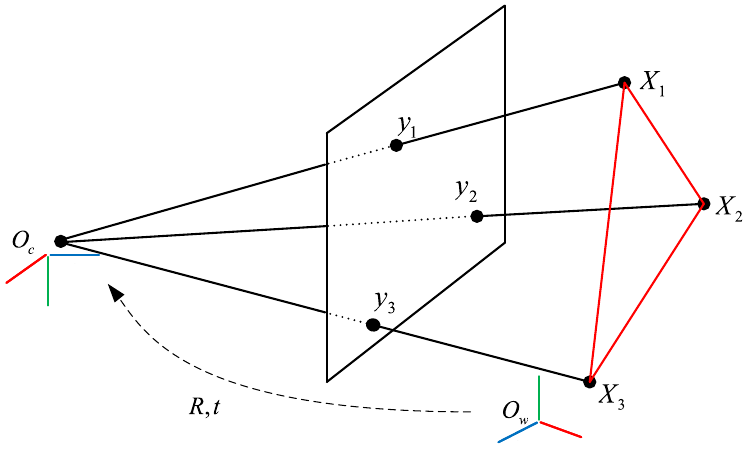}
   \caption{The perspective-three-point problem}
   \label{fig:camera_model}
\end{figure}

% For a pinhole camera model, consider three 3D points XXX in the world coordinate system. Their projection points on the normalized image plane are denoted as YYY. The coordinates of these projection points, normalized to have a unit length, are represented as mmm, with ∣m∣=1

For a pinhole camera model, consider three 3D points $\mathbf{X}_i=\left(x_i, y_i, z_i\right) , i \in\{1,2,3\}$ in the world coordinate system, as shown in the \cref{fig:camera_model}. Their projection points on the normalized image plane are represented as $\mathbf{y}_i =\left(u_i, v_i, 1\right), i \in\{1,2,3\}$. The coordinates of $\mathbf{y}_i$, after being normalized to have a unit norm, are represented as 
$\mathbf{m}_i \in \mathbb{R}^3$, with $\left|\mathbf{m}_i\right|=1, i \in\{1,2,3\}$. The rigid transformation between these corresponding sets of points, $\mathbf{X}_i$ and $\mathbf{m}_i$, are as follows:
\begin{equation}
\begin{aligned}
d_i \mathbf{m}_i=\mathbf{R} \mathbf{X}_i+\mathbf{t},
\label{eq:transformation}
\end{aligned}
\end{equation}
where $d_i \in \mathbb{R}^{+}, d_i=\left|\mathbf{O}_c \mathbf{X}_i\right|, i \in\{1,2,3\}$, a positive real number, represents the distance from the 3D point $\mathbf{X}_i$ to the camera center $\mathbf{O}_c$; $\mathbf{R} \in S O(3)$ represents the rotation; and $\mathbf{t} \in \mathbb{R}^3$ represents the translation, $\mathbf{R}$ and $\mathbf{t}$ together define the pose of the camera.

% \begin{equation}
% \begin{aligned}
% \mathbf{m}_i=\frac{\mathbf{y}_i}{\left\|\mathbf{y}_i\right\|},\left\|\mathbf{y}_i\right\|=\sqrt{u_i^2+v_i^2+1}
% \label{eq:transformation}
% \end{aligned}
% \end{equation}

Following a similar strategy as in~\cite{Ding_2023_CVPR, Persson_2018_ECCV}, we first derive the equations to formulate the P3P problem as finding the intersection of two conics. Before recovering $\mathbf{R}$ and $\mathbf{t}$, we first calculate $d_i$ by taking pairwise differences of the three equations in~\eqref{eq:transformation} and then square both sides of each resulting equation to eliminate $\mathbf{R}$ and $\mathbf{t}$. 
Consequently, we derive the following equations:
\begin{equation}
\begin{aligned}
& d_i \mathbf{m}_i-d_j \mathbf{m}_j=\mathbf{R} \mathbf{X}_i-\mathbf{R} \mathbf{X}_j, \Rightarrow \\
& \left|d_i \mathbf{m}_i-d_j \mathbf{m}_j\right|^2=\left|\mathbf{R} \mathbf{X}_i-\mathbf{R} \mathbf{X}_j\right|^2, \Rightarrow \\
& d_i^2-2 d_i d_j \mathbf{m}_i^{\top} \mathbf{m}_j+d_j^2=\left|\mathbf{X}_i-\mathbf{X}_j\right|^2,
\end{aligned}
\label{eq:distances}
\end{equation}
where \( \left|\mathbf{m}_i\right|=1 \), which implies that \( \mathbf{m}_i^T \mathbf{m}_i = 1 \). Similarly, for \( \mathbf{m}_j \), it is also confirmed that \( \mathbf{m}_j^T \mathbf{m}_j = 1 \). 

% To express the three equations in \eqref{eq:distances} in a more specific form, we have
We can express the three equations in~\eqref{eq:distances} as:
\begin{equation}
\begin{aligned}
& d_1^2-2 d_1 d_2 \mathbf{m}_1^{\top} \mathbf{m}_2+d_2^2=\left|\mathbf{X}_1-\mathbf{X}_2\right|^2, \\
& d_1^2-2 d_1 d_3 \mathbf{m}_1^{\top} \mathbf{m}_3+d_3^2=\left|\mathbf{X}_1-\mathbf{X}_3\right|^2, \\
& d_2^2-2 d_2 d_3 \mathbf{m}_2^{\top} \mathbf{m}_3+d_3^2=\left|\mathbf{X}_2-\mathbf{X}_3\right|^2.
\end{aligned}
\label{eq:cosines}
\end{equation}

% To solve the three equations for $d_1$, $d_2$, and $d_3$, we perform the following change of variables: 
We solve the above system of equations by the following change of variables:
\begin{equation}
\begin{aligned}
x=d_1 / d_3,\ y=d_2 / d_3,
\end{aligned}
\label{eq:var_change}
\end{equation} where \(d_1\), \(d_2\), and \(d_3\) are positive real numbers, making \(x\) and \(y\) positive real values. We then replace $d_1$, $d_2$, and $d_3$ in~\eqref{eq:cosines} with $x$ and $y$ by dividing the first two equations by the third equation, which yields two quadratic equations in the variables $x$ and $y$:
\begin{equation}
\begin{aligned}
x^2-2 m_{12} x y+(1-a) y^2+2 a m_{23} y-a & =0,
\end{aligned}
\label{eq:conics1}
\end{equation}
\begin{equation}
\begin{aligned}
x^2-b y^2-2 m_{13} x+2 b m_{23} y+1-b & =0,
\end{aligned}
\label{eq:conics2}
\end{equation}
where
\begin{equation}
\begin{aligned}
& a=\left|\mathbf{X}_1-\mathbf{X}_2\right|^2/\left|\mathbf{X}_2-\mathbf{X}_3\right|^2, \\
& b=\left|\mathbf{X}_1-\mathbf{X}_3\right|^2/\left|\mathbf{X}_2-\mathbf{X}_3\right|^2, \\ 
& m_{12}=\mathbf{m}_1^{\top} \mathbf{m}_2, m_{13}=\mathbf{m}_1^{\top} \mathbf{m}_3, m_{23}=\mathbf{m}_2^{\top} \mathbf{m}_3 .
\label{eq:fractions}
\end{aligned}
\end{equation}

Our next step is to find the positive real solutions of the two quadratic equations, which correspond to the intersection points of two conics in the positive real domain. These solutions will subsequently be used to determine the values of $d_i$. Once the values of $d_i$ have been obtained, we will proceed to compute the rotation matrix $\mathbf{R}$ and the translation vector $\mathbf{t}$.

\section{Our method}
Using homogeneous coordinates, the two quadratic equations (\ref{eq:conics1}) and (\ref{eq:conics2}) can be reformulated as the following matrix representations
\begin{equation}
\begin{aligned}
{{\mathbf{x}}^T}{\mathbf{C}_1\mathbf{x}} = 0,
\label{eq:matrices1_form}
\end{aligned}
\end{equation}
\begin{equation}
\begin{aligned}
{{\mathbf{x}}^T}{\mathbf{C}_2\mathbf{x}} = 0,
\label{eq:matrices2_form}
\end{aligned}
\end{equation}
% \begin{equation}
% \begin{aligned}
% {{\mathbf{x}}^T}\left[ {\begin{array}{@{}ccc@{}}
%   1&{ - {m_{12}}}&0 \\ 
%   { - {m_{12}}}&{1 - a}&{a{m_{23}}} \\ 
%   0&{a{m_{23}}}&{ - a} 
% \end{array}} \right]{\mathbf{x}} = {{\mathbf{x}}^T}{\mathbf{C}_1\mathbf{x}} = 0,
% \label{eq:matrices1_form}
% \end{aligned}
% \end{equation}
% \begin{equation}
% \begin{aligned}
% {{\mathbf{x}}^T}\left[ {\begin{array}{@{}ccc@{}}
%   1&0&{ - {m_{13}}} \\ 
%   0&{ - b}&{b{m_{23}}} \\ 
%   { - {m_{13}}}&{b{m_{23}}}&{1 - b} 
% \end{array}} \right]{\mathbf{x}} = {{\mathbf{x}}^T}{\mathbf{C}_2\mathbf{x}} = 0,
% \label{eq:matrices2_form}
% \end{aligned}
% \end{equation}
where $\mathbf{x}=\left[\begin{array}{@{}c@{}}
x,y,1\end{array}\right]^T$ , and the matrices $\mathbf{C}_1$ and $ \mathbf{C}_2$ are defined as
\begin{equation}
\begin{aligned}
\mathbf{C}_1\propto\left[\begin{array}{@{}ccc@{}}
1 & -m_{12} & 0 \\
-m_{12} & 1-a & a m_{23} \\
0 & a m_{23} & -a
\end{array}\right], \\
% \label{eq:matrices1}
% \end{aligned}
% \end{equation}
% \begin{equation}
% \begin{aligned}
\mathbf{C}_2\propto\left[\begin{array}{@{}ccc@{}}
1 & 0 & -m_{13} \\
0 & -b & b m_{23} \\
-m_{13} & b m_{23} & 1-b
\end{array}\right].
\label{eq:matrices2}
\end{aligned}
\end{equation}

We follow a similar strategy as in~\cite{mancini2024intersection} to transform these conics into a new coordinate system, simplifying them and finding their intersections in this new coordinate system.

Let $\mathbf{x} \propto \mathbf{H}\mathbf{x}^{\prime}$, where \(\mathbf{H}\) represents a homography matrix, \(\mathbf{x}\) is a point in the original coordinate system, and \(\mathbf{x}^{\prime}\) is the corresponding point in the new coordinate system. Using this transformation, the equations of the conics in the original coordinate system are transformed into the new coordinate system as follows:
\begin{equation}
\begin{aligned}
 {{\mathbf{(Hx^{\prime})}}^T}{\mathbf{C}_1\mathbf{Hx^{\prime}}} = 0 \quad \Rightarrow \quad {{\mathbf{x^{\prime}}}^T}{\mathbf{C}_1^{\prime}\mathbf{x^{\prime}}} = 0,
\label{eq:xc1x}
\end{aligned}
\end{equation}
\begin{equation}
\begin{aligned}
 {{\mathbf{(Hx^{\prime})}}^T}{\mathbf{C}_2\mathbf{Hx^{\prime}}} = 0 \quad \Rightarrow \quad {{\mathbf{x^{\prime}}}^T}{\mathbf{C}_2^{\prime}\mathbf{x^{\prime}}} = 0.
\label{eq:xc2x}
\end{aligned}
\end{equation}
Here, \(\mathbf{C}_1^{\prime}\) and \(\mathbf{C}_2^{\prime}\) are the transformed conic matrices in the new coordinate system, defined as:
\begin{equation}
\begin{aligned}
& {\mathbf{C}_1^{\prime}\propto\mathbf{H}^T \mathbf{C}_1 \mathbf{H}}, \\
& {\mathbf{C}_2^{\prime}\propto\mathbf{H}^T \mathbf{C}_2 \mathbf{H}}.
\label{eq:C1prime}
\end{aligned}
\end{equation}

\subsection{Selecting three points on the first conic}
Before determining the transformation matrix, we first need to select three points $\mathbf{p}_1$, $\mathbf{p}_2$, and $\mathbf{p}_3$ on the first conic to serve as reference points for its calculation.

% Unlike the generalized point-selection approach in \cite{mancini2024intersection}, which does not consider the situation of introducing complex numbers, the special property of the conic in the P3P problem allow us to choose points that avoid the need for complex arithmetic, thereby improving computational efficiency.
The generalized point-selection approach in~\cite{mancini2024intersection} does not account for the potential introduction of complex numbers, which increases computational complexity. In contrast, the unique properties of the conics in the P3P problem offer a better strategy, allowing us to select points that inherently avoid the need for complex arithmetic.
% Unlike the generalized point-selection approach in~\cite{mancini2024intersection}, which does not take into account the potential introduction of complex numbers, leading to increased computational complexity, the unique properties of the conics in the P3P problem provide us a better strategy to choose points that inherently avoid the need for complex arithmetic.
By selecting points that lie entirely within the real domain, we not only simplify the mathematical operations involved but also reduce the risk of errors, thereby greatly improving computational efficiency and stability.
% Unlike the approach in \cite{mancini2024intersection}, which can introduce complex numbers and increase computational complexity, the unique properties of the conics in the P3P problem provide us a better strategy to select points entirely within the real domain. This strategy avoids complex arithmetic, simplifies operations, and improves computational efficiency and stability.

% From \eqref{eq:fractions}, since the ratio of the squared distances is positive, it follows that $a > 0$. Therefore, for the conic \eqref{eq:conics1}, the line $y = 0$ must intersect the conic at two points, with the homogeneous coordinates of the intersection points $\mathbf{p}_2$ and $\mathbf{p}_3$ given by:
% From \eqref{eq:fractions}, since the ratio of the squared distances is positive, it follows that $a > 0$. For the conic \eqref{eq:conics1}, the line $y = 0$ must intersect the conic at two distinct points within the real domain，因为将y=0代入到圆锥曲线方程，可以得到$x^2=a$，其中a大于0，The homogeneous coordinates of the intersection points, denoted as $\mathbf{p}_2$ and $\mathbf{p}_3$.
% From \eqref{eq:fractions}, since the ratio of the squared distances is positive, it follows that $a > 0$. For the conic \eqref{eq:conics1}, the line $y = 0$ must intersect the conic at two distinct points within the real domain, because substituting $y = 0$ into the conic equation yields $x^2 = a$, where $a > 0$. The homogeneous coordinates of the intersection points, denoted as $\mathbf{p}_2$ and $\mathbf{p}_3$, are therefore real and distinct.
We begin by selecting the second and third points, \(\mathbf{p}_2\) and \(\mathbf{p}_3\). From (\ref{eq:fractions}), it follows that $a > 0$ because the ratio of the squared distances is positive.
For the conic (\ref{eq:conics1}), when the line $y = 0$ is substituted into the conic equation, we obtain a simplified equation of the form $x^2 = a$, where $a$ is a positive constant. This implies that the conic intersects the line $y = 0$ at two distinct points in the real domain, specifically at $x = \sqrt{a}$ and $x = -\sqrt{a}$. The homogeneous coordinates of these two points, denoted as $\mathbf{p}_2$ and $\mathbf{p}_3$, are then expressed as:

\begin{equation}
\mathbf{p}_{2,3} \propto\left[\begin{array}{@{}c@{}}
\pm \sqrt{a} \\
0 \\
1
\label{eq:p2_p3}
\end{array}\right].
\end{equation}

% \begin{equation}
% \begin{aligned}
% \mathbf{p}_1 \propto\left[\begin{array}{@{}c@{}}
% \sqrt{a} \\
% \frac{2 m_{12} \sqrt{a}-2 a m_{23}}{1-a} \\
% 1
% \end{array}\right],
% \mathbf{p}_{2,3} \propto\left[\begin{array}{@{}c@{}}
% \pm \sqrt{a} \\
% 0 \\
% 1
% \label{eq:p1,p2,p3}
% \end{array}\right]
% \end{aligned}
% \end{equation}

\begin{figure}[h]
  \centering
   \includegraphics[width=0.85\linewidth]{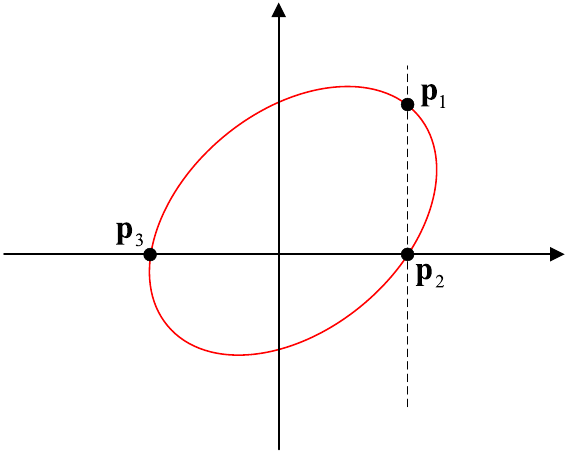}
   \caption{Three points selection on a conic}
   \label{fig:conic1}
\end{figure}

For the selection of the first point $\mathbf{p}_1$, as shown in \cref{fig:conic1}, we can find the equation of the line perpendicular to the \(x\)-axis and passing through $\mathbf{p}_2$ as $x = \sqrt{a}$. The intersection of this line with the conic provides the first point $\mathbf{p}_1$, with the coordinates given as follows:
\begin{equation}
\begin{aligned}
\mathbf{p}_1 \propto\left[\begin{array}{@{}c@{}}
\sqrt{a} \\
\frac{2 m_{12} \sqrt{a}-2 a m_{23}}{1-a} \\
1
\end{array}\right].
\label{eq:p1}
\end{aligned}
\end{equation}

% % 使用figure*环境实现跨栏
% \begin{figure*}[h!]
%     \centering
%     % subfigure (a)
%     \begin{subfigure}[b]{0.23\textwidth}
%         \centering
%         \includegraphics[width=\textwidth]{./figure/hyperbola.pdf}
%         \caption{Selection on a Hyperbola}
%         \label{fig:hyperbola_points}
%     \end{subfigure}
%     \hfill
%     % subfigure (b)
%     \begin{subfigure}[b]{0.23\textwidth}
%         \centering
%         \includegraphics[width=\textwidth]{./figure/ellipse.pdf}
%         \caption{Selection on an Ellipse}
%         \label{fig:ellipse_points}
%     \end{subfigure}
%     \hfill
%     % subfigure (c)
%     \begin{subfigure}[b]{0.23\textwidth}
%         \centering
%         \includegraphics[width=\textwidth]{./figure/parabola.pdf}
%         \caption{Selection on a Parabola}
%         \label{fig:parabola_points}
%     \end{subfigure}

%     \caption{Three points selection on conics in special cases.}
%     \label{fig:conic_three_cases}
% \end{figure*}

\begin{figure}[h!]
    \centering
    % subfigure (a)
    \begin{subfigure}[b]{0.23\textwidth}
        \centering
        \includegraphics[width=\textwidth]{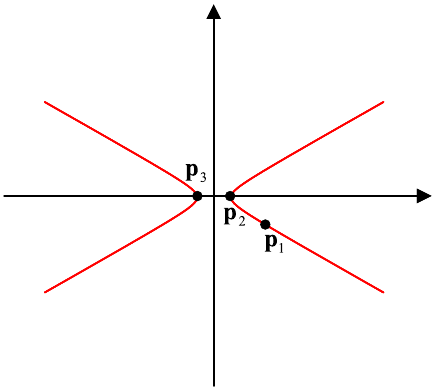}
        \caption{Selection on a Hyperbola}
        \label{fig:hyperbola_points}
    \end{subfigure}
    \hfill
    % subfigure (b)
    \begin{subfigure}[b]{0.23\textwidth}
        \centering
        \includegraphics[width=\textwidth]{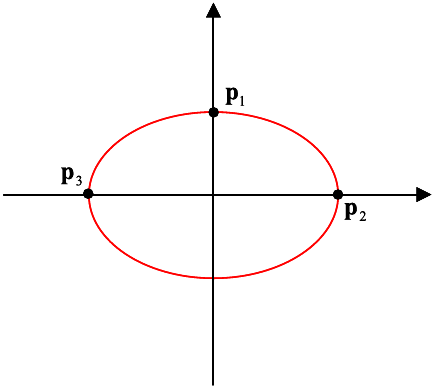}
        \caption{Selection on an Ellipse}
        \label{fig:ellipse_points}
    \end{subfigure}
    \hfill
    % subfigure (c)
    \begin{subfigure}[b]{0.23\textwidth}
        \centering
        \includegraphics[width=\textwidth]{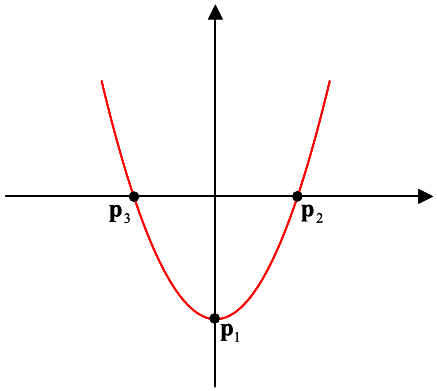}
        \caption{Selection on a Parabola}
        \label{fig:parabola_points}
    \end{subfigure}    
    \caption{Three points selection on conics in special cases.}
    \label{fig:conic_three_cases}
\end{figure}

% This situation occurs because when \(a = 1\), the distances \(\left|\mathbf{X}_1 - \mathbf{X}_2\right|\) and \(\left|\mathbf{X}_2 - \mathbf{X}_3\right|\) are equal, which geometrically corresponds to a special case causes the denominator to become zero, which is mathematically problematic and must be avoided. 
% When selecting the third point in the way, two special cases must be considered. First, when $a = 1$, the second element of $\mathbf{p}_1$ becomes undefined due to division by zero, resulting in an undefined or infinite value. Second, if the denominator of the second element of $\mathbf{p}_1$ is zero, $\mathbf{p}_1$ coincides with $\mathbf{p}_2$, indicating that both \(\mathbf{p}_1\) and \(\mathbf{p}_2\) represent the same point. 
When selecting the first point \( \mathbf{p}_1 \) in this way, one special case must be carefully considered to ensure the robustness of the point selection process. When \( a = 1 \), the second element of \( \mathbf{p}_1 \) becomes undefined due to division by zero. This occurs because, in this case, the distances \( \left|\mathbf{X}_1 - \mathbf{X}_2\right| \) and \( \left|\mathbf{X}_2 - \mathbf{X}_3\right| \) become equal, resulting in a special geometric configuration where the conic equation may simplify or degenerate, causing the denominator of the second element of \( \mathbf{p}_1 \) to approach zero, which is mathematically problematic. If the denominator of the second element of \( \mathbf{p}_1 \) becomes zero or too small, it implies that \( \mathbf{p}_1 \) coincides with or is too close to \( \mathbf{p}_2 \), making them indistinguishable as distinct points on the conic. This leads to a reduction in the number of unique points available for constructing the transformation matrix, potentially causing numerical inaccuracies. To address this, a threshold is set in the implementation to filter out cases where \(\mathbf{p}_1\) and \(\mathbf{p}_2\) are either coincident or too close, ensuring the stability of the computation.

To address this issue, we employ the following approach to select the first point \( \mathbf{p}_1 \), avoiding cases where \( \mathbf{p}_1 \) and \( \mathbf{p}_2 \) are either coincident or too close, ensuring robustness and accuracy.
We classify the conic \( \mathbf{C}_1 \) based on its discriminant, \(\Delta = 4\left(m_{12}^2 + a - 1\right)\), which allows us to identify it as one of three types, as shown in \cref{fig:conic_three_cases}. 

If \(\Delta > 0\), \( \mathbf{C}_1 \) is a hyperbola, as shown in \cref{fig:hyperbola_points}. We select the point whose \( x \)-coordinate is \( 1 \) greater than the \( x \)-coordinate of \( \mathbf{p}_2 \). This selection ensures that \(\mathbf{p}_1\) is distinct from \(\mathbf{p}_2\) and avoids issues that may arise from coincident points or from selecting a point too close to \(\mathbf{p}_2\). 

If \(\Delta < 0\), \( \mathbf{C}_1 \) is an ellipse, as shown in \cref{fig:ellipse_points}. We select the point with the x-coordinate in the halfway between \(\mathbf{p}_2\) and \(\mathbf{p}_3\) (i.e., \(x=0\)) on the conic. 
This choice is motivated by the fact that \(\mathbf{p}_1\) is the furthest point from both \(\mathbf{p}_2\) and \(\mathbf{p}_3\), making it as distinct as possible.

If \( \Delta = 0 \), \( \mathbf{C}_1 \) is a parabola, as shown in \cref{fig:parabola_points}. We use the same point selection strategy as in the case where \( \Delta < 0 \).

% We first determine the type of the conic \(\mathbf{C}_1\) using its discriminant, \(\Delta = 4\left(m_{12}^2 + a - 1\right)\), which allows us to classify the conic into one of three categories, as illustrated in \cref{fig:conic_three_cases}. Specifically, if \(\Delta > 0\), it is a hyperbola; if \(\Delta < 0\), \(\mathbf{C}_1\) is an ellipse; and if \(\Delta = 0\), it is a parabola.

% For the hyperbola \(\mathbf{C}_1\), as shown in \cref{fig:hyperbola_points}, we select the nearest integer \(x\)-coordinate that is greater than the \(x\)-coordinate of \(\mathbf{p}_2\). 
% selecting the point corresponding to the average of the \(x\)-coordinates of \(\mathbf{p}_2\) and \(\mathbf{p}_3\) guarantees that \(\mathbf{p}_1\) lies on the ellipse or parabola \(\mathbf{C}_1\), ensuring that \(\mathbf{p}_1\) is both a valid point on the curve and distinct from \(\mathbf{p}_2\) and \(\mathbf{p}_3\).

% If a=1,,,,. %需要重新选择
% 求p1和p2对应的极线，我们求两条极线交点可以得到第四个点p0

\subsection{Finding a proper homography matrix}
Once \(\mathbf{p}_1\), \(\mathbf{p}_2\), and \(\mathbf{p}_3\) have been properly selected, the next step is finding the homography matrix $\mathbf{H}$ that transforms the original coordinates to the new coordinate system~\cite{mancini2024intersection}. To do this, we first calculate the polar lines of $\mathbf{p}_1$ and $\mathbf{p}_2$  with respect to the conic \(\mathbf{C}_1\). The two polar lines can be computed as $\mathbf{l}_1 \propto \mathbf{C}_1 \mathbf{p}_1$ and $\mathbf{l}_2 \propto \mathbf{C}_1 \mathbf{p}_2$. Let \(\mathbf{p}_0\) be the intersection point of the two polar lines
\begin{equation}
\mathbf{p}_0 \propto \mathbf{l}_1 \times \mathbf{l}_2.
\label{eq:p0}
\end{equation}
Since $\mathbf{p}_1$ and $\mathbf{p}_2$ are real points, $\mathbf{l}_1$ and $\mathbf{l}_2$ with respect to the conic $\mathbf{C}_1$ are also real. Therefore, $\mathbf{p}_0$ is a real point as well.

Now we have four points in the original coordinates system, with no three points being collinear. 
% $\mathbf{e}_0, \mathbf{e}_1, \mathbf{e}_2$, and $\mathbf{e}_3$.
We can find a non-singular linear transformation that maps four points $\mathbf{e}_0^T\propto[1,0,0], \mathbf{e}_1^T\propto[0,1,0]$, $\mathbf{e}_2^T\propto[0,0,1]$ and $\mathbf{e}_3^T\propto[1,1,1]$ in the new coordinate system to the four points $\mathbf{p}_0, \mathbf{p}_1, \mathbf{p}_2$, and $\mathbf{p}_3$ in the original coordinate system.
% We can find a non-singular linear transformation that maps the conic $\mathbf{C}_1$ in the original coordinate system defined by the reference points $\mathbf{p}_0, \mathbf{p}_1$, and $\mathbf{p}_2$ to the conic $\mathbf{C}_1^{\prime}$ in new coordinate system defined by the canonical basis with the reference points $\mathbf{e}_0^T=[1,0,0], \mathbf{e}_1^T=[0,1,0]$, and $\mathbf{e}_2^T=[0,0,1]$.
The mapping is represented as the homography matrix $\mathbf{H}$, and the transformation of these points can be expressed as follows:
\begin{equation}
\begin{aligned}
\mathbf{H} \mathbf{e}_i=\lambda_i \mathbf{p}_i, 
\label{eq:Hei}
\end{aligned}
\end{equation}
where $\lambda_i$ represents a non-zero scale factor, $i=0, \ldots, 3$.
We can select the first three equations from~\eqref{eq:Hei} (i.e., for $i=0,1,2$) to construct $\mathbf{H}$
\begin{equation}
\begin{aligned}
\mathbf{H} \propto\left[\begin{array}{@{}ccc@{}}
\lambda_0 \mathbf{p}_0 & \lambda_1 \mathbf{p}_1 & \lambda_2 \mathbf{p}_2
\end{array}\right].
\label{eq:H}
\end{aligned}
\end{equation}

Furthermore, we can fix the scaling by setting $\lambda_3=1$ and utilize the fourth equation of~\eqref{eq:Hei} (i.e., for $i=3$) to determine the values of $\lambda_0$, $\lambda_1$, and $\lambda_2$ as follows:

\begin{equation}
\begin{aligned}
& \left[\begin{array}{@{}ccc@{}}
\lambda_0 \mathbf{p}_0 & \lambda_1 \mathbf{p}_1 & \lambda_2 \mathbf{p}_2
\end{array}\right]\left[\begin{array}{@{}c@{}}
1 \\
1 \\
1
\end{array}\right]=\mathbf{p}_3,\Rightarrow \\
& \left[\begin{array}{@{}ccc@{}}
\mathbf{p}_0 & \mathbf{p}_1 & \mathbf{p}_2
\end{array}\right]\left[\begin{array}{@{}c@{}}
\lambda_0 \\
\lambda_1 \\
\lambda_2 
\end{array}\right]=\mathbf{p}_3.
\label{eq:solve_lambda}
\end{aligned}
\end{equation}

This matrix $\left[\begin{array}{@{}ccc@{}}
\mathbf{p}_0 & \mathbf{p}_1 & \mathbf{p}_2
\end{array}\right]$ in~\eqref{eq:solve_lambda} is invertible since the three points are not collinear. The values of $\lambda_0$, $\lambda_1$, and $\lambda_2$ can be determined by multiplying the second equation in~\eqref{eq:solve_lambda} by the inverse of $\left[\begin{array}{@{}ccc@{}}
\mathbf{p}_0 & \mathbf{p}_1 & \mathbf{p}_2
\end{array}\right]$, which can then be substituted into~\eqref{eq:H} to obtain $\mathbf{H}$.

For any given point $\mathbf{x}^{\prime}$ in the new coordinate system, we get the corresponding point in the original coordinate system using
\begin{equation}
\begin{aligned}
\mathbf{x} \propto \mathbf{H} \mathbf{x}^{\prime}.
\label{eq:anypoint}
\end{aligned}
\end{equation}
% so, we just substitute \cref{{eq:anypoint}} into \eqref{eq:matrices1_form} and \eqref{eq:matrices2_form} get \

\subsection{Finding the intersection points}
Given the transformation in~\eqref{eq:anypoint}, the two conics in the new coordinate system can be expressed as shown in~\eqref{eq:C1prime}. The corresponding transformed conic matrices $\mathbf{C}_1^{\prime}$ and $\mathbf{C}_2^{\prime}$ are given by~\cite{mancini2024intersection}
\begin{equation}
\begin{aligned}
\mathbf{C}_1^{\prime} \propto \mathbf{H}^T \mathbf{C}_1 \mathbf{H} \propto\left[\begin{array}{@{}ccc@{}}
a_1^{\prime} & b_1^{\prime} / 2 & d_1^{\prime} / 2 \\
b_1^{\prime} / 2 & c_1^{\prime} & e_1^{\prime} / 2 \\
d_1^{\prime} / 2 & e_1^{\prime} / 2 & f_1^{\prime}
\end{array}\right],
\label{eq:C1_prime}
\end{aligned}
\end{equation}
\begin{equation}
\begin{aligned}
\mathbf{C}_2^{\prime} \propto \mathbf{H}^T \mathbf{C}_2 \mathbf{H} \propto\left[\begin{array}{@{}ccc@{}}
a_2^{\prime} & b_2^{\prime} / 2 & d_2^{\prime} / 2 \\
b_2^{\prime} / 2 & c_2^{\prime} & e_2^{\prime} / 2 \\
d_2^{\prime} / 2 & e_2^{\prime} / 2 & f_2^{\prime}
\end{array}\right].
\label{eq:C2_prime_abcdef}
\end{aligned}
\end{equation}

By using the matrix \(\mathbf{H}\) defined in~\eqref{eq:H}, \(\mathbf{C}_1^{\prime}\) can be further written as:
\begin{equation}
\begin{aligned}
\mathbf{C}_1^{\prime} \propto \left[\begin{array}{@{}ccc@{}}
2 & 0 & 0 \\
0 & 0 & -1 \\
0 & -1 & 0
\end{array}\right].
\label{eq:parabola}
\end{aligned}
\end{equation}

This result can be demonstrated by examining the transformation $\mathbf{H}^T \mathbf{C}_1 \mathbf{H}$, which maps the conic $\mathbf{C}_1$ in the coordinate system defined by the reference points $\mathbf{p}_0$, $\mathbf{p}_1$, and $\mathbf{p}_2$ to the conic $\mathbf{C}_1^{\prime}$ in another coordinate system defined by the canonical basis, as shown in~\eqref{eq:C1_prime}, with the reference points $\mathbf{e}_0^T, \mathbf{e}_1^T$, and $\mathbf{e}_2^T$. 
Since $\mathbf{p}_0$ is the pole of the line through $\mathbf{p}_1$ and $\mathbf{p}_2$ in the first coordinate system, $\mathbf{e}_0$ is the pole of the line through $\mathbf{e}_1$ and $\mathbf{e}_2$ in the second (canonical) coordinate system. The point $\mathbf{e}_1$ is at the infinity along the y-axis, and the point $\mathbf{e}_2$ is at the origin. Therefore, the line passing through $\mathbf{e}_1$ and $\mathbf{e}_2$ is the vertical line $x=0$, which can be expressed in homogeneous coordinates as $\mathbf{u}_{0}^T=[1,0,0]$. The relationship between this line and the pole $\mathbf{e}_0$, which is a point at the infinity along the x-axis, satisfies the equation $\mathbf{u}_{0} \propto \mathbf{C}'_{1} \mathbf{e}_{0}$. Hence, the parameters $b'_1$ and $d'_1$ in $\mathbf{C}'_{1}$ must be zeros. Moreover, the reference points $\mathbf{e}_1$, $\mathbf{e}_2$, and $\mathbf{e}_{3}$ are points on the conic $\mathbf{C}'_{1}$, and thus $\mathbf{e}^T_{i} \mathbf{C}'_{1} \mathbf{e}_{i} = 0$, where $i=1,2,$ and $3$. This gives us the following constraints: $c'_1=0$, $f'_1=0$, and $a'_1+b'_1+c'_1+d'_1+e'_1+f'_1 = 0$. When we combine all these constraints, we get $a'_1 + e'_1 = 0$, which corresponds to the matrix in~\eqref{eq:parabola}, representing a parabola.

 By substituting~\eqref{eq:parabola} into~\eqref{eq:xc1x} and~\eqref{eq:C2_prime_abcdef} into~\eqref{eq:xc2x}, two new conics are obtained
\begin{equation}
\begin{aligned}
\mathbf{x}^{\prime T} \mathbf{C}_1^{\prime} \mathbf{x}^{\prime}=0 \quad \Rightarrow \quad x^{\prime 2}=y^{\prime},
\label{eq:new_equation1}
\end{aligned}
\end{equation}
\begin{equation}
\begin{aligned}
\mathbf{x}^{\prime T} \mathbf{C}_2^{\prime} \mathbf{x}^{\prime}=0 \quad \Rightarrow \quad & a_2^{\prime} x^{\prime 2}+b_2^{\prime} x^{\prime} y^{\prime}+c_2^{\prime} y^{\prime 2} \\
& +d_2^{\prime} x^{\prime}+e_2^{\prime} y^{\prime}+f_2^{\prime}=0.
\label{eq:new_equation2}
\end{aligned}
\end{equation}
% \begin{multline}
% \mathbf{x}'^{T} \mathbf{C}'_2 \mathbf{x}' = 0 \Rightarrow a'_2 x'^2 + b'_2 x' y' + c'_2 y'^2 \\
% + d'_2 x' + e'_2 y' + f'_2 = 0
% \label{eq:new_equation2}
% \end{multline}

Next, by substituting~\eqref{eq:new_equation1} into~\eqref{eq:new_equation2} and rearranging the terms, gives a simple quartic equation with only five terms: 
\begin{equation}
\begin{aligned}
c_2^{\prime} x^{\prime 4}+b_2^{\prime} x^{\prime 3}+\left(a_2^{\prime}+e_2^{\prime}\right) x^{\prime 2}+d_2^{\prime} x^{\prime}+f_2^{\prime}=0.
\label{eq:quartic}
\end{aligned}
\end{equation}

The solution of~\eqref{eq:quartic} yields up to four real roots \(x^{\prime}_j\), \(j \in [1, N]\), where \(N\leq 4\), which can be solved using Ferrari's method as described in~\cite{cardano2007rules}, with the detailed solution formulas provided in the Supplementary Material (Section 1). 
% These solutions, when substituted into \eqref{eq:new_equation1} to obtain the corresponding $y^{\prime}_j$, provide up to four corresponding intersection points \((x^{\prime}_j, y^{\prime}_j)\) in the new coordinates.
Substituting these solutions into~\eqref{eq:new_equation1} yields the corresponding values of \(y^{\prime}_j\), providing up to four intersection points \((x^{\prime}_j, y^{\prime}_j)\) in the new coordinate system.

Since $\mathbf{p}_0$, $\mathbf{p}_1$, $\mathbf{p}_2$, and $\mathbf{p}_3$ are all real points, the coefficients $\lambda_0$, $\lambda_1$, and $\lambda_2$ from (\ref{eq:solve_lambda}) must also be real. This confirms that the homography matrix $\mathbf{H}$ in (\ref{eq:H}) is real. Given that $x$ and $y$ represent ratios of distances, $\mathbf{x}$ in the original coordinate system is real. According to~\eqref{eq:anypoint}, $\mathbf{x}^{\prime}$ in the new coordinate system and the corresponding points $(x^{\prime}_j, y^{\prime}_j)$ are also real.

By using the matrix $\mathbf{H}$, the points $\left(x_j^{\prime}, y_j^{\prime}\right)$ in the new coordinate system are transformed into $\left(x_j, y_j\right)$ in the original coordinate system as described in~\eqref{eq:anypoint}. Since the points in the original coordinate system have positive coordinates, this condition can be used to filter the correct solutions. This ultimately identifies the intersection points of the two conics in the original coordinate system.

\subsection{Recovering R and t}
In the following step, we use the intersection points of the two conics to solve for the unknowns \(d_1\), \(d_2\), and \(d_3\)~\cite{Ding_2023_CVPR, Persson_2018_ECCV}. Since two conics can have up to four intersection points, each intersection point corresponds to a distinct set of \(d_i\) values. Given the relationships \(x = d_1 / d_3\) and \(y = d_2 / d_3\), we can substitute these expressions into any equation in~\eqref{eq:cosines}, first solving for \(d_3\), and then using this result to determine \(d_1\) and \(d_2\). For each set of \(d_i\) values obtained, we refine them using the Gauss-Newton optimization method, a technique that has also been employed in several other works, including~\cite{ke2017efficient, nakano2019simple, Persson_2018_ECCV, Ding_2023_CVPR}.

From the first equation in~\eqref{eq:distances}, we can derive the following relationship:
\begin{equation}
\begin{aligned}
& \left[\begin{array}{@{}ccc@{}}
d_1 \mathbf{m}_1-d_2 \mathbf{m}_2 &  d_1 \mathbf{m}_1-d_3 \mathbf{m}_3 & d_2 \mathbf{m}_2-d_3 \mathbf{m}_3\end{array}\right] \\
& =\mathbf{R} \left[\begin{array}{@{}ccc@{}}
\mathbf{X}_1-\mathbf{X}_2 &  \mathbf{X}_1-\mathbf{X}_3 & \mathbf{X}_2-\mathbf{X}_3 \end{array}\right].
\end{aligned}
\label{eq:rotation}
\end{equation}

We observe that the three vectors $\mathbf{X}_1-\mathbf{X}_2$, $\mathbf{X}_1-\mathbf{X}_3$ and $\mathbf{X}_2-\mathbf{X}_3$ in~\eqref{eq:rotation} are coplanar. As a result, the matrix they form is rank-deficient and therefore noninvertible, which means we cannot directly use its inverse to solve for $\mathbf{R}$. To overcome this limitation, we introduce a vector perpendicular to this plane, $(\mathbf{X}_1-\mathbf{X}_2) \times (\mathbf{X}_1-\mathbf{X}_3)$, thereby increasing the rank of the matrix and making it invertible. This adjustment allows us to solve for $\mathbf{R}$:
\begin{equation}
\begin{aligned}
\mathbf{Y}=\mathbf{R} \mathbf{X} \quad \Rightarrow \quad \mathbf{R}=\mathbf{Y} \mathbf{X}^{-1},
\end{aligned}
\label{eq:rotation1}
\end{equation}
where
\begin{equation}
\begin{aligned}
& \mathbf{Y} = \left[\begin{array}{@{}ccc@{}}
d_1 \mathbf{m}_1-d_2 \mathbf{m}_2 &  d_1 \mathbf{m}_1-d_3 \mathbf{m}_3 & \mathbf{n}_y\end{array}\right], \\
& \mathbf{X} = \left[\begin{array}{@{}ccc@{}}
\mathbf{X}_1-\mathbf{X}_2 &  \mathbf{X}_1-\mathbf{X}_3 & \mathbf{n}_x \end{array}\right], \\
& \mathbf{n}_x = (\mathbf{X}_1-\mathbf{X}_2) \times (\mathbf{X}_1-\mathbf{X}_3), \\
& \mathbf{n}_y = (d_1 \mathbf{m}_1-d_2 \mathbf{m}_2) \times (d_1 \mathbf{m}_1-d_3 \mathbf{m}_3).
\end{aligned}
\label{eq:rotation2}
\end{equation}

Once $\mathbf{R}$ is determined, we can substitute it into any of the three equations in~\eqref{eq:transformation} to solve for $\mathbf{t}$:
\begin{equation}
\begin{aligned}
\mathbf{t} = d_i \mathbf{m}_i - \mathbf{R} \mathbf{X}_i.
\label{eq:translation}
\end{aligned}
\end{equation}

Each set of \(d_i\) values will produce one pose $(\mathbf{R}_i, \mathbf{t}_i)$. The complete procedure is summarised in Algorithm 1.

\begin{algorithm}
\caption{Conic Transformation Approach to P3P}\label{alg:rotation_translation}
\textbf{Input:} 3D points \( \mathbf{X}_i \), normalized image points \( \mathbf{y}_i \), \( i = 1, 2, 3 \) \\
\textbf{Output:} Poses \(\lbrace \mathbf{R}_j, \mathbf{t}_j \rbrace_{j=1,\dots,N}\), where \(N \leq 4\)
\begin{algorithmic}[1]
\State Normalize \( \mathbf{y}_i \) to unit norm \( \mathbf{m}_i = \mathbf{y}_i / \|\mathbf{y}_i\| \)
\State Compute \( a, b, m_{12}, m_{13}, m_{23} \) based on~\eqref{eq:fractions}
\State Construct matrices \( \mathbf{C}_1,\) and \(\mathbf{C}_2 \) in~\eqref{eq:matrices1_form},~\eqref{eq:matrices2_form} and~\eqref{eq:matrices2} using homogeneous coordinates
\State Find three points \( \mathbf{p}_1 \),\( \mathbf{p}_2 \), and \( \mathbf{p}_3 \) on the first conic
\State Compute the intersection point \(\mathbf{p}_0\) of the polar lines corresponding to \(\mathbf{p}_1\) and \(\mathbf{p}_2\) using~\eqref{eq:p0}
\State Solve for three \(\lambda_k\), \( k = 0, 1, 2 \) using~\eqref{eq:solve_lambda}
\State Compute the transformation matrix \(\mathbf{H}\) using~\eqref{eq:H}
\State Transform \(\mathbf{C}_1\) and \(\mathbf{C}_2\) to the new coordinate system using~\eqref{eq:C1_prime} and ~\eqref{eq:C2_prime_abcdef}, yielding \(\mathbf{C}_1^{\prime}\) and \(\mathbf{C}_2^{\prime}\)
% \State Compute \(\mathbf{C}_1^{\prime}\) in the new coordinate system using \eqref{eq:C2_prime_abcdef}
% \State Compute the real root \(x^{\prime}_j\) of the quartic equation given by the coefficients in \eqref{eq:quartic} using Ferrari's method
\State Solve the quartic equation~\eqref{eq:quartic} using Ferrari's method~\cite{cardano2007rules} to find the real roots \(x^{\prime}_j, j\in[1,N]\)
\For{\(j=1,\dots,N\) }
    \State Compute the intersection points \((x^{\prime}_j, y^{\prime}_j)\).
    \State Transform $\left(x_j^{\prime}, y_j^{\prime}\right)$ back to the original coordinates \phantom{using~\eqref{eq:anypoint}} \hspace{-1.1cm} using~\eqref{eq:anypoint}
    \State Filter the positive $\left(x_j, y_j\right)$ values
    \State Rewrite~\eqref{eq:var_change}, as \(d_1 = x d_3\) and \(d_2 = y d_3\)
    % and \phantom{\eqref{eq:cosines}} \hspace{-0.6cm} \eqref{eq:cosines}
    \State Substitute \(d_1\) and \(d_2\) into~\eqref{eq:cosines} and use one of the \phantom{using~\eqref{eq:anypoint}} \hspace{-1.05cm} equations to solve for \(d_3\)
    \State Refine \(d_i\) using Gauss-Newton~\cite{ke2017efficient, nakano2019simple, Persson_2018_ECCV, Ding_2023_CVPR}
    \State Compute \(\mathbf{R}_j\) and \(\mathbf{t}_j\) using~\eqref{eq:rotation1} and~\eqref{eq:rotation2}
\EndFor
\end{algorithmic}
\end{algorithm}

\section{Experiments}
In this section, we compare our proposed method with several state-of-the-art solvers in terms of numerical stability and runtime efficiency. To make a fair comparison, all solvers are implemented in C++ and run on a desktop computer with an AMD Ryzen 5 5600GE 3.4GHz CPU. Additionally, they were evaluated using the same random synthetic data without noise, applying consistent criteria for determining correct solutions, as described in~\cite{Ding_2023_CVPR}. 

The proposed method is compared with the following P3P solvers: the state-of-the-art P3P solver by Ding \textit{et al.}~\cite{Ding_2023_CVPR}, the solver by Persson and Nordberg~\cite{Persson_2018_ECCV}, the solver by Nakano~\cite{nakano2019simple}, the solver by Ke \textit{et al.}~\cite{ke2017efficient} and the solver by Kneip \textit{et al.}~\cite{kneip2011novel}. These solvers represent a diverse set of approaches, providing a comprehensive benchmark for assessing the effectiveness of our method. Two important considerations should be noted. 
Firstly, for Nakano's method, since only the MATLAB implementation is available online, we reimplemented it in C++ to ensure consistency in our comparisons.
% Firstly, for Nakano's method, only the MATLAB implementation is available online; therefore, to ensure consistency in our comparisons, we utilized the C++ implementation provided by Ding \textit{et al.}~\cite{Ding_2023_CVPR} for this comparison. 
Secondly, regarding Ding's method, there are two implementations in GitHub published in November 2023 together with~\cite{Ding_2023_CVPR} and a more recent implementation published in April 2024.
To ensure that our comparison was comprehensive and accounted for any potential improvements or bug fixes, we conducted experiments using both the original implementation and the updated version.

The synthetic data used in this paper is generated following the approach in~\cite{Persson_2018_ECCV}, which is consistent with the method used in~\cite{Ding_2023_CVPR}. To give a general overview, the synthetic data consists of random rigid transformations and randomly distributed observations. For the random rigid transformations, the ground truth rotation matrix \( \mathbf{R}_{gt} \) is obtained by converting a unit quaternion drawn from an isotropic Gaussian distribution, and the ground truth translation vector \(\mathbf{t}_{gt} \) is generated from a standard normal distribution. 
For the randomly distributed observations, the normalized image points \(\mathbf{y}_i =\left(u_i, v_i\right)\) are first generated by a uniform sampling 2D coordinates within the range \([-1, 1]\). The corresponding 3D points \( \mathbf{X}_i \) are then computed using \( \mathbf{R}_{gt} \) and \( \mathbf{t}_{gt} \) using the formula \( \mathbf{X}_i = \mathbf{R}_{gt}^\top (d_i \mathbf{m}_i - \mathbf{t}_{gt}) \), where the depth \( d_i \) is a uniform sampled as a random positive value within the interval \([0.1, 10]\). Additionally, it is important to note that cases where the three normalized image points or their corresponding 3D points are collinear are excluded, as such configurations lack sufficient geometric constraints to determine the transformation parameters. However, near-degenerate data are retained to assess the robustness of the algorithms, ensuring they can handle challenging scenarios effectively.

\subsection{Numerical stability}

% \begin{figure}[h]
%     \centering
%     % subfigure (a)
%     \begin{subfigure}[b]{0.235\textwidth}
%         \centering
%         \includegraphics[width=\textwidth]{./figure/mygraph_R.eps}
%         \caption{Logarithm of rotation error}
%         \label{fig:error_R}
%     \end{subfigure}
%     \hfill
%     % subfigure (b)
%     \begin{subfigure}[b]{0.235\textwidth}
%         \centering
%         \includegraphics[width=\textwidth]{./figure/mygraph_t.eps}
%         \caption{Logarithm of translation error}
%         \label{fig:error_t}
%     \end{subfigure}
%     \hfill
%     % subfigure (c)
%     \begin{subfigure}[b]{0.235\textwidth}
%         \centering
%         \includegraphics[width=\textwidth]{./figure/mygraph.eps}
        % \caption{Logarithm of the sum of rotation and translation errors}
        % \label{fig:error_R_t}
%     \end{subfigure}    
%     % \caption{Gaussian kernel smoothed histograms of a logarithmic sum of rotation and translation errors across various algorithms for 100,000 runs on noise-free data.}
%     \caption{Gaussian kernel smoothed histograms of the logarithmic rotation, translation, and combined errors for various algorithms over 100,000 runs on noise-free data.}
%     \label{fig:error}
% \end{figure}

\begin{figure}[h]
  \centering
  \includegraphics[width=0.9\linewidth,trim={0.cm 0cm 0cm 0cm},clip]{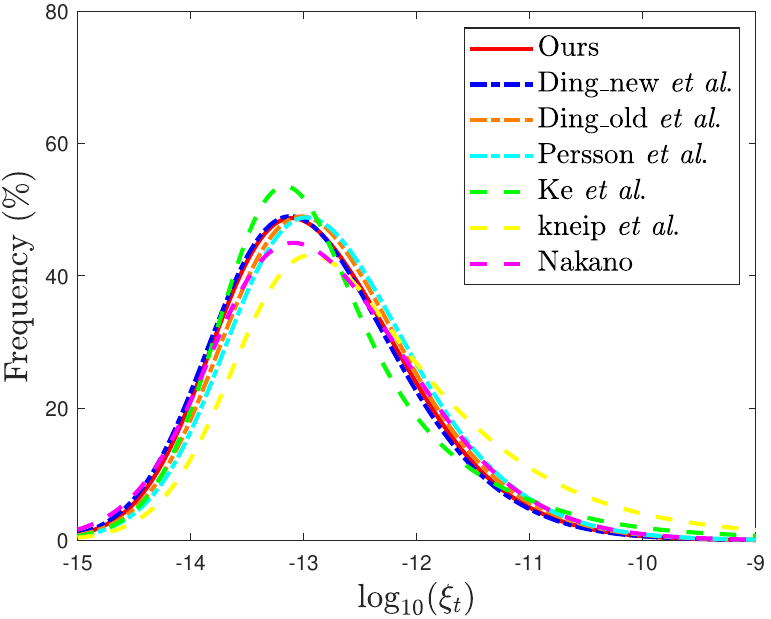}
   \caption{Gaussian kernel smoothed histograms of a logarithmic sum of rotation and translation errors across various algorithms for 100,000 runs on noise-free data.}
   \label{fig:error}
\end{figure}

The numerical stability is determined by the error between the solutions obtained by the algorithm for each data and the ground truth. The error consists of the rotation error and translation error, which are defined as \( \xi_R = \|\mathbf{R}_{gt} - \mathbf{R}\|_{L1} \) and \( \xi_t = \|\mathbf{t}_{gt} - \mathbf{t}\|_{L1} \), respectively.
\cref{fig:error} shows the Gaussian kernel smoothed histograms of the logarithm (base 10) of the sum of the rotation and translation errors, represented as \( \log_{10}(\xi_R + \xi_t) \), for various algorithms, highlighting the frequency distribution of errors across 100,000 runs on noise-free data.
From \cref{fig:error}, it can be observed that both our method and the other methods are numerically stable. For separate visualizations of the rotation and translation errors, see the Supplementary Material (Section 2).
% \cref{fig:error} shows the Gaussian kernel smoothed histograms of the logarithmic rotation error (\cref{fig:error_R}), translation error (\cref{fig:error_t}), and combined error (\cref{fig:error_R_t}) for various algorithms, highlighting the frequency distribution of errors across 100,000 runs on noise-free data.  
% From \cref{fig:error}, it can be observed that both our method and the other methods are numerically stable.

Additionally, we computed three error metrics, namely mean, median, and max, on data of size \(10^7\), as shown in the \cref{tab:error_comparison}.  A sample is defined as a failure case if \( \xi_R + \xi_t > 10^{-6} \). To ensure a fair comparison, all failure cases from all solvers have been removed from this analysis. It can be seen that our proposed method achieves the smallest mean error, the solver by Ke \textit{et al.}~\cite{ke2017efficient} performs best in terms of median error, and the solver by Ding \textit{et al.}~\cite{Ding_2023_CVPR} shows the best performance for max error. The random sampling seed was set to 1 in the corresponding implementation.

\begin{table}[t]
    \centering
    \begin{tabular}{@{}cccc@{}}
        \toprule
        Method & Mean & Median & Max  \\
        \midrule
        Ours & \textbf{3.907e-12} & 1.244e-13   & 9.311e-7 \\
        Ding\_new \textit{et al.}~\cite{Ding_2023_CVPR} & 3.999e-12 & 1.173e-13  &  \textbf{4.951e-7} \\
        Ding\_old \textit{et al.}~\cite{Ding_2023_CVPR} & 5.595e-12 & 1.452e-13   & 9.861e-7 \\
        Persson \textit{et al.}~\cite{Persson_2018_ECCV} &6.025e-12 & 1.613e-13 & 9.549e-7 \\
        Ke \textit{et al.}~\cite{ke2017efficient} & 2.287e-10 & \textbf{1.09e-13} & 9.998e-7 \\
        kneip \textit{et al.}~\cite{kneip2011novel} & 6.265e-10  & 2.523e-13  & 9.999e-7 \\
        Nakano~\cite{nakano2019simple}&  7.91e-12 & 1.371e-13 & 8.21e-7 \\
        \bottomrule
    \end{tabular}
    \caption{Comparison of the mean, median, and max values of the errors with the current state-of-the-art solvers. The best results are highlighted in bold.}
    \label{tab:error_comparison}
\end{table}

% In Figure\cref{fig:error}, 显示的是不同方法在100,000大小的noise-free数据上的误差曲线图，其中横坐标是误差值是由旋转误差和平移误差相加取对数得到，误差值estimation和groud truth之间误差一范数，纵坐标为误差出现的频率。all the errors all the methods are numerically stable. 我们的方法和丁的方法误差曲线相近，轻微差一些，但好于其他的方法。Ke的方法误差集中在-13处，但误差较大处也不能收敛，例如-10处的曲线值大于前两者方法。其余方法曲线整体靠右，误差值偏大。在（ding在新的代码中，更新了三种方法，我们选择与三种中精度最高速度最快的方法比较）

\subsection{Solution discussion}
\begin{table*}[t]
    \centering
    % \begin{tabular}{@{}l@{\hskip 1pt}c@{\hskip 1pt}c@{\hskip 1pt}c@{\hskip 1pt}c@{\hskip 1pt}c@{\hskip 1pt}c@{\hskip 1pt}c@{}}
    \begin{tabular}{@{}cccccccc@{}}
        \toprule
        Method & Ours 
        & \multicolumn{1}{c}{\begin{tabular}[c]{@{}c@{}}Ding\_new\\ \textit{et al.}~\cite{Ding_2023_CVPR}\end{tabular}} 
        &\multicolumn{1}{c}{\begin{tabular}[c]{@{}c@{}}Ding\_old\\ \textit{et al.}~\cite{Ding_2023_CVPR}\end{tabular}}
        &\multicolumn{1}{c}{\begin{tabular}[c]{@{}c@{}}Persson\\ \textit{et al.}~\cite{Persson_2018_ECCV}\end{tabular}}
        &\multicolumn{1}{c}{\begin{tabular}[c]{@{}c@{}}Ke\\ \textit{et al.}~\cite{ke2017efficient}\end{tabular}}
        & \multicolumn{1}{c}{\begin{tabular}[c]{@{}c@{}}Kneip\\ \textit{et al.}~\cite{kneip2011novel}\end{tabular}}
        & \multicolumn{1}{c}{\begin{tabular}[c]{@{}c@{}}Nakano\\~\cite{nakano2019simple}\end{tabular}}\\
        \midrule
        
        Valid solutions & 16824038 & 16824040 & 16824032 & 16824039 & 17385099 & 24147180 & 16823171 \\
        Unique solutions & 16824038 & 16824040 & 16824028 & 16824035 & 16849138 & 16826291 & 16823110 \\
        Duplicates &   0        & 0        & 0        & 0        & 161322   & 3011     & 8\\
        Good solutions & 9999999  & 10000000  & 9999995 & 9999997  & 9999639  & 9999677  & 9999382 \\
        No solution & 1        & 0        & 5        & 3        & 361      & 323      & 618  \\
        Ground truth & 9999997  & 9999998  & 9999990  & 9999990  & 9997158  & 9990970  & 9999306 \\
        Incorrect solutions & 0        & 0        & 4        & 4        & 374639   & 7317878  & 53\\
        \bottomrule
    \end{tabular}
    \caption{Solution comparison with the current state-of-the-art solvers on \(10^7\) samples.}
    \label{tab:sol_comparison}
\end{table*}

\begin{table*}[h]
    \centering
    \begin{tabular}{@{}cccccccc@{}}
        \toprule
        Timing (ns) & Ours 
        & \multicolumn{1}{c}{\begin{tabular}[c]{@{}c@{}}Ding\_new\\ \textit{et al.}~\cite{Ding_2023_CVPR}\end{tabular}} 
        &\multicolumn{1}{c}{\begin{tabular}[c]{@{}c@{}}Ding\_old\\ \textit{et al.}~\cite{Ding_2023_CVPR}\end{tabular}}
        &\multicolumn{1}{c}{\begin{tabular}[c]{@{}c@{}}Persson\\ \textit{et al.}~\cite{Persson_2018_ECCV}\end{tabular}}
        &\multicolumn{1}{c}{\begin{tabular}[c]{@{}c@{}}Ke\\ \textit{et al.}~\cite{ke2017efficient}\end{tabular}}
        & \multicolumn{1}{c}{\begin{tabular}[c]{@{}c@{}}Kneip\\ \textit{et al.}~\cite{kneip2011novel}\end{tabular}}
        & \multicolumn{1}{c}{\begin{tabular}[c]{@{}c@{}}Nakano\\~\cite{nakano2019simple}\end{tabular}}\\
        \midrule
        Mean   & \textbf{190.3} & 199.7  & 203.1  & 264.9  & 375.0  & 647.0  & 759.9  \\
        Median & \textbf{190.2} & 199.6  & 202.9  & 264.9  & 374.8  & 646.6  & 759.3  \\
        Min    & \textbf{189.8} & 199.2  & 202.5  & 264.4  & 373.8  & 645.6  & 757.3  \\
        Max    & \textbf{195.0} & 203.6  & 207.0  & 265.9  & 381.4  & 664.3  & 776.2  \\
        \midrule
        Speed up & \textbf{1.0494}       & 1.0    & 0.9833 & 0.7539 & 0.5325 & 0.3087 & 0.2628  \\
        \bottomrule
    \end{tabular}
    \caption{Running times comparison averaged over $(10^{7})$ trials with 100 times each.}
    \label{tab:time_comparison}
\end{table*}

% We further analyzed the solutions provided by our method and the current state-of-the-art solvers on \(10^7\) samples and the results are shown in \cref{tab:sol_comparison}. This table presents several evaluation metrics. Below, we explain these metrics in the context of a single sample. The values in the table represent the cumulative results obtained across all \(10^7\) samples.
We further analyzed the solutions provided by our method and the current state-of-the-art solvers on \(10^7\) samples, and the results are shown in \cref{tab:sol_comparison}. This table presents several evaluation metrics, with the values representing the cumulative results obtained across all \(10^7\) samples.

Valid solutions refer to the number of pose solutions \((\mathbf{R}, \mathbf{t})\) generated by the solver. Unique solutions refer to the number of those valid solutions that meet all of the following five conditions, as specified in \cite{Ding_2023_CVPR}, with each condition adhering to the same thresholds: 1) \(\left| \operatorname{det}(\mathbf{R} \mathbf{R}^T) - 1 \right| < 10^{-6}\); 2) \(\left| \operatorname{det}(\mathbf{R}) - 1 \right| < 10^{-6}\); 3) A quaternion norm corresponding to \( \mathbf{R} \), with \( |\|\mathbf{q}\| - 1| < 10^{-5} \); 4) a reprojection error of the three 3D points, projected from the world coordinate system into the camera coordinate system, of less than \(10^{-4}\); 5) any duplicates among these solutions are identified and removed, ensuring that only distinct solutions remain.
% "Valid solutions" refer to the total number of pose solutions \((\mathbf{R}, \mathbf{t})\) generated by the solver. 
% "Unique solutions" refer to the number of those valid solutions that meet all of the following five conditions: 1) \( \operatorname{det}(\mathbf{R} \mathbf{R}^T) < 10^{-6} \); 2) \( \operatorname{det}(\mathbf{R}) < 10^{-6} \); 3) a reprojection error of three 3D points, projected from the world coordinate system into the camera coordinate system, of less than \(10^{-4}\); 4) and a quaternion norm corresponding to \(\mathbf{R}\) of less than \(10^{-5}\); 5) If multiple solutions are generated, duplicates are identified and removed. 
For each sample, if there is at least one unique solution, we refer to this case as a good solution. Good solutions refer to the number of such cases.
Duplicates refer to the number of cases where two solutions from the same trial satisfy \( \xi_R + \xi_t < 10^{-5} \).
No solution refers to the number of cases where neither unique solutions nor duplicates are found for a trial.
Ground truth refers to the number of samples that have at least one solution satisfying \( \xi_R + \xi_t < 10^{-6} \).
Incorrect solutions refer to the number of cases where valid solutions are obtained, but none meet the criteria for unique solutions or duplicates.

From \cref{tab:sol_comparison}, it can be seen that for \(10^7\) samples, the proposed method effectively finds good solutions and the ground truth. While it slightly underperforms compared to Ding\_new in terms of overall metrics, it outperforms other methods.
Compared to the methods of Ding \textit{et al.}~\cite{Ding_2023_CVPR} and Persson \textit{et al.}~\cite{Persson_2018_ECCV}, which solve the P3P problem by finding the roots of a cubic equation, our method not only outperforms the methods of Persson and Ding\_old in terms of ground truth and good solution metrics, but it also results in fewer incorrect solutions and fewer instances where no solution is found.
Compared with the methods of Ke \textit{et al.}~\cite{ke2017efficient} and Kneip \textit{et al.}~\cite{kneip2011novel}, which solve the P3P problem by finding the roots of a quartic equation like ours, the results show that many incorrect or duplicate solutions are obtained. This is because they solve for all the roots of the quartic equation, omitting the imaginary part to obtain four solutions, which leads to inaccuracies and reduced computational efficiency.
Additionally, the method of Nakano \textit{et al.}~\cite{nakano2019simple}, which also solves the quartic equation, filters out roots with small imaginary parts by setting a threshold $(10^{-8})$, extracting the real part as the real root. The setting of this threshold can influence the accuracy to a certain degree. A threshold set too low may filter out fewer correct roots, while one set too high may introduce more incorrect roots.
In our proposed method, when solving quartic equations, we address potential complex number introduction during square root operations by first evaluating whether the values under the square root are greater than or equal to zero. If the values are negative, we discard those cases, thereby avoiding the need for complex number computations.
% In our proposed method, when solving quartic equations and encountering situations where square root operations might introduce complex numbers, we evaluate and discard any complex roots to avoid the introduction of complex number computations.
The results in the table correspond to a random sampling seed that was also set to 1 in the corresponding implementation.

\subsection{Execution time}
We tested the proposed method against other current state-of-the-art solvers on \(10^7\) samples and also made a comparison of execution times, as shown in \cref{tab:time_comparison}.
For each solver, we ran tests on a dataset of \(10^7\) samples, with each sample being processed 100 times. From the table, it is evident that the proposed method is computationally more efficient than other methods, being faster in terms of mean, median, minimum, and maximum times.
The proposed method is about 4.8\% faster than the current state-of-the-art solver by Ding \textit{et al.}~\cite{Ding_2023_CVPR}.

The main reasons for the speed improvement are as follows: First, we only solve for the real roots of the quartic equation and incorporate a check during the square root process within the function. Second, by constructing a transformation matrix, we transform one of the conics into a standard parabola. Due to the simplicity of the parabolic form, which is easier to handle mathematically, inserting it into the conic equation after the second transformation simplifies it, making it faster to obtain the corresponding coefficients for the quartic equation. Third, when acquiring the coordinate transformation matrix, we leverage the characteristic of quadratic curves in the P3P problem, which always intersects the line $y=0$, allowing us to find two real points on the conic curve quickly. Using this property to further locate a third real point prevents the emergence of complex points and thus prevents the occurrence of complex transformation matrices. The proposed approach effectively reduces the computational burden associated with handling complex numbers and simplifies the derivation of the coefficients in the quartic equation, leading to a more streamlined and efficient solution.

\subsection{Discussion}
Since the intersection coordinates of the conics represent depth ratios, they are positive real numbers. In our method, after applying the transformation, we primarily utilize the constraint that the values are real numbers. However, we recognize that there could be additional constraints within the quartic equation that could further refine our solution process, particularly constraints ensuring that the solutions of the quartic equation correspond to positive values before the transformation. While our present approach focuses on the most straightforward real-number constraint, we acknowledge the potential to explore and integrate these additional positive constraints to further enhance both accuracy and computational efficiency.
Besides, future efforts are planned to focus on applying the solver to practical data within real-world pipelines.
\section{Conclusion}
% In this paper, we propose a novel P3P method based on a conic transformation approach. We begin by converting the P3P problem into a problem of finding the real intersection points of two conics. To achieve this, we construct a transformation matrix and transform these conics into a new coordinate system, where one of them is a standard parabola. This transformation simplifies the problem, allowing us to derive a quartic equation with a low number of terms and easily obtainable coefficients, thereby reducing the original P3P problem to solving the real roots of this quartic equation. Furthermore, by avoiding the introduction of complex numbers both in the solution of the quartic equation and in the construction of the transformation matrix, we improve the computational efficiency of the method. Extensive experiments demonstrate that our method is faster than other methods, while being on par with the state-of-the-art in terms of stability and robustness. The implementation of the method will be published upon acceptance of the paper.
In this paper, we study the P3P problem by revisiting the known problem of finding the intersection of two conics. Our contributions to this approach are two-fold. First, we propose a coordinate system transformation that converts one of the conics into a standard parabola. This transformation allows us to express the intersection of the conics as a quartic equation in the new coordinate system, with coefficients that can be quickly computed. 
The second contribution is a strategy to avoid introducing a complex transformation matrix and solving for complex solutions of the quartic equation, via a clever selection of three points on one conic. 
Our approach improves the efficiency of the P3P solver by avoiding computing with complex numbers. Extensive experiments demonstrate that our method is faster than other methods while being on par with the state-of-the-art in terms of stability and robustness. 
% The implementation of our method will be published upon acceptance of the paper.
The implementation of our method is available at: \url{https://github.com/hayden-86/p3p-solver}.
\section{Acknowledgements}
This work was supported by Research Council of Finland under the project "Methods and Applications for High-Efficiency Polynomial Solvers" (grant no. 355970).

% \section*{Appendix}

%%%%%%%%% REFERENCES
{\small
\bibliographystyle{ieee_fullname}
\bibliography{egbib}
}

\end{document}